\documentclass[conference]{IEEEtran}
\usepackage{algorithm}
\usepackage{algorithmic}

\usepackage{ifpdf}

\ifCLASSINFOpdf
   \usepackage[pdftex]{graphicx}
\else
\fi
\begin{document}
%
\title{Band Selection and Classification of  Hyperspectral Images using Mutual Information: 
An algorithm based on minimizing the error probability using the inequality of Fano}

\author{
		\IEEEauthorblockN{Elkebir Sarhrouni *}
		\IEEEauthorblockA{LRIT, FSR, UMV-A\\Rabat, Morocco \\
		Email: sarhrouni436@yahoo.frl}
\and
		\IEEEauthorblockN{Ahmed Hammouch}
		\IEEEauthorblockA{LRIT, FSR, UMV-A\\
		LRGE, ENSET, UMV-SOUISSI\\
		Rabat, Morocco \\
		Email: hammouch\_a@yahoo.com}
\and
		\IEEEauthorblockN{Driss Aboutajdine}
		\IEEEauthorblockA{LRIT, FSR, UMV-A\\Rabat, Morocco \\
		Email: aboutaj@fsr.ac.ma}}


%


\maketitle

	\begin{abstract}
	 Hyperspectral image is a substitution of more than a hundred images, called bands, of the same region. They are taken at juxtaposed frequencies. The reference image of the region is called Ground Truth map (GT). the problematic is how to find the good bands to classify the pixels of regions; because the bands can be not only redundant, but a source of confusion, and decreasing so the accuracy of classification. Some methods use Mutual Information (MI) and threshold, to select relevant bands. Recently there’s an algorithm selection based on mutual information, using bandwidth rejection and a threshold to control and eliminate redundancy. The band top ranking the MI is selected, and if its neighbors have sensibly the same MI with the GT, they will be considered redundant and so discarded. This is the most inconvenient of this method, because this avoids the advantage of hyperspectral images:: some precious information can be discarded. In this paper we’ll make difference between useful and useless redundancy. A band contains useful redundancy if it contributes to decreasing error probability. According to this scheme, we introduce new algorithm using also mutual information, but it retains only the bands minimizing the error probability of classification. To control redundancy, we introduce a complementary threshold. So the good band candidate must contribute to decrease the last error probability augmented by the threshold. This process is a wrapper strategy; it gets high performance of classification accuracy but it is expensive than filter strategy.
	\end{abstract}


%
\IEEEpeerreviewmaketitle

\section{Introduction}
In the feature classification domain, the choice of data affects widely the results. For the Hyperspectral image, the bands don’t all contain the information; some bands are irrelevant like those affected by various atmospheric effects, see Figure.4, and decrease the classification accuracy. And there exist redundant bands to complicate the learning system and product incorrect prediction [14]. Even the bands contain enough information about the scene they may can’t predict the classes correctly if the dimension of space images, see Figure.3, is so large that needs many cases to detect the relationship between the bands and the  scene (Hughes phenomenon) [10]. We can reduce the dimensionality of hyperspectral images by selecting only the relevant bands (feature selection or subset selection methodology), or extracting, from the original bands, new bands containing the maximal information about the classes, using any functions, logical or numerical (feature extraction methodology) [11][9]. Here we focus on the feature selection using mutual information. Hyperspectral images have three advantages regarding the multispectral images [6], see Figure.1\\ \\
\textbf{First:} the hyperspectral image contains more than a hundred images but the multispectral contains three at ten images.\\
\textbf{Second:} hyperspectral image has a spectral resolution (the central wavelength divided by de width of spectral band) about a hundred, but that of multispectral is about ten.\\
\textbf{Third:} the bands of a hyperspectral image is regularly spaced, those of multispectral image is large and irregularly spaced.\\ 
\textbf{Comment:} when we reduce hyperspectral images dimensionality, we must save the precision and high discrimination of substances given by hyperspectral image.\\


\begin{figure}[!h]
\centering
\includegraphics[width=3in]{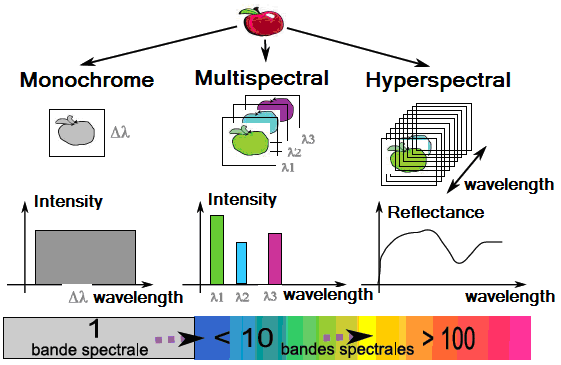}
\caption{Precision an dicrimination added by hyperspectral images}
\label{fig_sim}
\end{figure}

In this paper we use the Hyperspectral image AVIRIS 92AV3C (Airborne Visible Infrared Imaging Spectrometer). [2]. It contains 220 images taken on the region "Indiana Pine" at "north-western Indiana", USA [1]. The 220 called bands are taken between 0.4µm and 2.5µm. Each band has 145 lines and 145 columns. The ground truth map is also provided, but only 10366 pixels are labeled fro 1 to 16. Each label indicates one from 16 classes. The zeros indicate pixels how are not classified yet, Figure.2.\\

\begin{figure}[!th]
\centering
\includegraphics[width=3in]{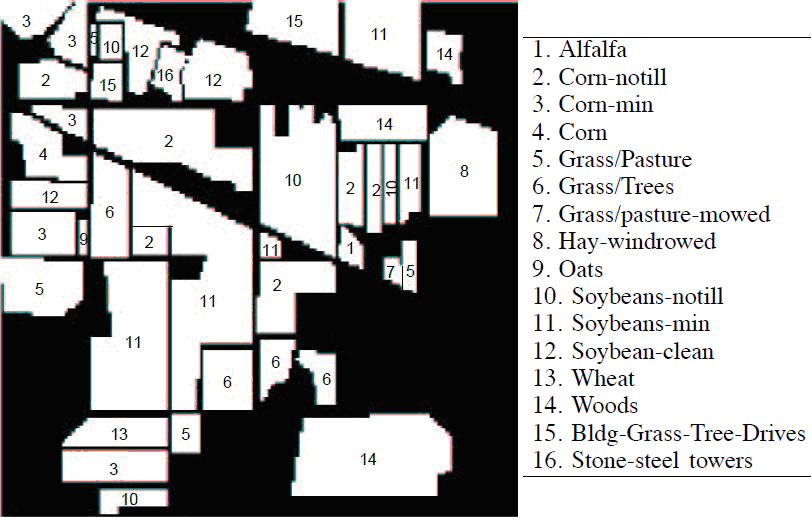}
\caption{The Ground Truth map of AVIRIS 92AV3C and the 16 classes }
\label{fig_sim}
\end{figure}

The hyperspectral image AVIRIS 92AV3C contains numbers between 955 and 9406. Each pixel of the ground truth map has a set of 220 numbers (measures) along the hyperspectral image. This numbers (measures) represent the reflectance of the pixel in each band. So the pixel is shown as vector off 220 components. 
Figure.3 shows the vector pixel’s notion [7]. So reducing dimensionality means selecting only the dimensions caring a lot of information regarding the classes.

\begin{figure}[!h]
\centering
\includegraphics[width=3in]{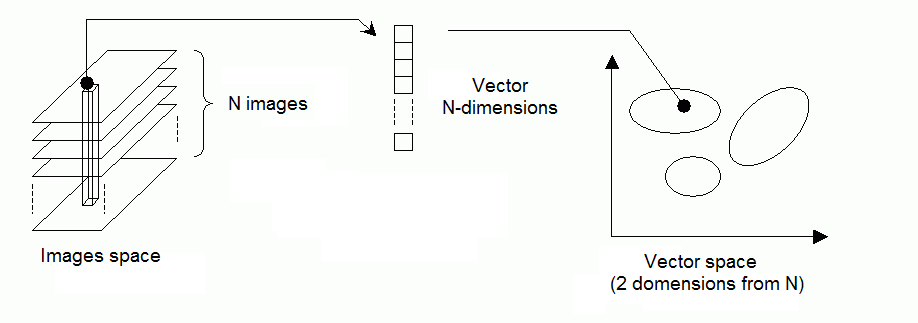}
\caption{The notion of  pixel vector }
\label{fig_sim}
\end{figure}

We can also note that not all classes are carrier of information. In Figure. 4, for example, we can show the effects of atmospheric affects on bands: 155, 220 and other bands. This hyperspectral image presents the problematic of dimensionality reduction.

\section{Mutual information based selection}

In this paragraph we inspect a recent method called band selection scheme using mutual information, and a rejection bandwidth algorithm to eliminate redundancy [3]. [7].

\subsection{Definition of mutual information}
This is a measure of exchanged information between tow ensembles of random variables A and B :
\[
I(A,B)=\sum\;log_2\;p(A,B)\;\frac{p(A,B)}{p(A).p(B)}
\]

Considering the ground truth map, and bands as ensembles of random variables, we calculate their interdependence. 

Geo [3] uses also the average of bands 170 to 210, to product an estimated ground truth map, and use it instead of the real truth map. Their curves are similar. Figure 4.

\subsection{Bands selection using mutual information}
From the mutual information curve, we can make threshold, and we retain the bands that have mutual information value above threshold. But the adjacent bands are possibly redundant. Geo[3] propose an algorithm to eliminate redundancy. It’s described in [3] as following: "Let \textit{B}$_{m}$  be the band that maximizes the mutual information. And N the number of \textit{B}$_{m}$  neighboring bands. We define: \[ d(n)=\Delta (MI(n)-MI(n-1)) \] If \textit{max}$_{n}$d(n) is down to a $threshold$ only, \textit{B}$_{m}$ is retained", i.e. its N neighbors will be discarded, because they my be redundante .
Let X be the number of bands to be selected. At some point in the selection process, let S be the set of selected bands, and let R be the set of remaining bands. We initialize the process with SS=Ø and R={1,2,,,220}.

\begin{algorithm}  
\vspace{0.20cm}                    
\caption{ Proposed  dimentionality Reduction and Redundancy control}  
        
\label{algo 1}  
\begin{algorithmic}
    \WHILE {$|SS| <  X$} 
        \STATE Select  \textit{band index}$_{s}$ $S$=\textit{argmax}$_{s}$  MI(s)
	\STATE  Neighbours set \textit{N}=$\{n|n=S-(B+1),…,S,…,S+B $\}

           \IF {$max_{s}d(n)< threshold$}
                \STATE $SS\gets \textit{SS} \cup \textit{S}$ and  $R\gets \textit{R} \setminus\textit{SS}\setminus\textit{N}$
	\ELSE 
	\STATE $SS\gets \textit{SS} \cup \textit{S}$ and  $R\gets \textit{R} \setminus\textit{SS}$
        \ENDIF
\ENDWHILE
\end{algorithmic}
\end{algorithm}

For more details refer to [3].

\subsection{Discussion and critics of method}

This algorithm is applied at mutual information calculated with the estimated ground truth map Figure.4. Like at [3] 50
The most inconvenient of this method is how it measures redundancy: small values of d(n)=Δ(MI(n)-MI(n-1)) doesn’t necessary an expression of redundancy. It’s seed at [3] that the neighboring bands are possibly redundant. So with this method, the advantage (the precision viewed at section I) of hyperspectral images regarding the multispectral images is avoid, because the precious information can be avoided. 

\subsection{Partial conclusion}

In this section we inspect the effectiveness of mutual information based selection for hyperspectral images. In the next step we use also the mutual information. 

One inconvenient if this filter approach is the time made to adjust manually the parameters. It can be expensive. But the most inconvenient is t hat reduce redundancy by eliminating the precision given by the notion of hyperspectral images. We propose now an algorithm avoiding only useless redundancy. We apply this rule: "If a band decreases the error probability, it will be retained even if it contains redundant information".

\begin{figure}[!h]
\centering
\includegraphics[width=3in]{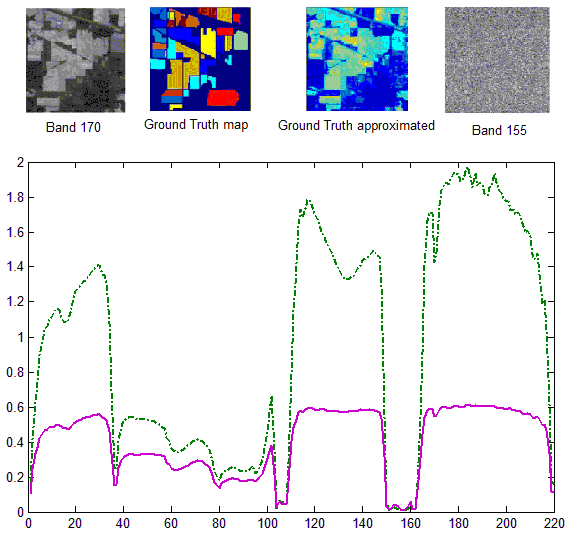}
\caption{Mutual information of AVIRIS  with the Ground Truth map (solid line) and with the ground apporoximated by averaging bands 170 to 210 (dashed line) .}
\label{fig_sim}
\end{figure}

\section{The mesure of error probability}
\subsection{Inequality of Fano}
Here we inspect the inequality of Fano [8]:
\[\;\frac{H(C/X)-1}{Log_2(N_c)}\leq\;P_e\leq\frac{H(C/X)}{Log_2}\; \]with :
\[
\;\frac{H(C/X)-1}{Log_2(N_c)}=\frac{H(C)-I(C;X)-1}{Log_2(N_c)}\; \] and :
\[    P_e\leq\frac{H(C)-I(C;X)}{Log_2}=\frac{H(C/X)}{Log_2}\; \]

The expression of conditional entropy\textit{ H(C/X)} is calculated between the ground truth map (i.e. the classes C) and the subset of bands candidates X. Nc is the number of classes. So when the features X have a higher value of mutual information with the ground truth map, (is more near to the ground truth map), the error probability will be lower. But it’s difficult to compute this conjoint mutual information\textit{ I(C;X)}, regarding the high dimensionality [14].This propriety makes Mutual Information a good criterion to measure resemblance between too bands, like it’s exploited in section II.
Furthermore, we will interest at case of one feature candidate X.\\
Corollary: for one feature X, as X approaches the ground truth map, the interval \textit{P}$_{e}$ is very small.

\subsection{Algorithm based on inequality of Fano }

Our idea is based on this observation: the band that has higher value of Mutual Information with the ground truth map can be a good approximation of it. So we note that the subset of selected bands are the good ones, if thy can generate an estimated reference map, sensibly equal the ground truth map. It’s clearly that’s an Incremental Wrapper-based Subset Selection (IWSS) approach[16] [13].

Our process of band selection will be as following: we order the bands according to value of its mutual information with the ground truth map. Then we initialize the selected bands ensemble with the band that has highest value of MI. At a point of process, we build an approximated reference map \textit{C}$_{est}$  with the already selected bands, and we put it instead of $X$ for computing the error probability (\textit{P}$_{e}$); the latest band added (at those already selected) must make \textit{P}$_{e}$ decreasing, if else it will be discarded from the ensemble retained.
Then we introduce a complementary threshold \textit{T}$_{h}$to control redundancy. So the band to be selected must make error probability less than ( \textit{P}$_{e}$ - \textit{T}$_{h}$) , where  \textit{P}$_{e}$ is calculated before adding it. The algorithm following shows more detail  of the process:

\textit{
Let SS be the ensemble of bands already selected and S the band candidate to be selected.  \textit{Build}$_{estimated}C()$ is a procedure to construct the estimated reference map. \textit{P}$_{e}$ is initialized with a value {\textit{P}$_{e}^*$ } . X the number of bands to be selected, $SS$ is empty and $R={1..220}$.}

\begin{algorithm}  
\vspace{0.20cm}                    
\caption{ Proposed for Dimentionality Reduction and Redundancy control}  
   
\label{algo 2:}  
\begin{algorithmic}
    \WHILE {$|SS| <  X$} 
        \STATE Select  \textit{band index}$_{s}$ $S$=\textit{argmax}$_{s}$  MI(s)
	\STATE $SS\gets \textit{SS} \cup \textit{S}$ and  $R\gets \textit{R} \setminus\textit{S}$
	\STATE  \textit{C}$_{est}$= \textit{Build}$_{estimated}C(SS)$
	\STATE  \[ Pe=\frac{H(C/C_{est})}{Log_2}\ - \frac{H(C/C_{est})-1}{Log_2(N_c)};\;\;\;\;\;\;\;\;\;\;\;\;\;\;\;\;\;\;\;\;\;\;\;\;\;\;																\;\;\;\;\;\;\;\;\;\;\;\;\;\;\;\;
														\;\;\;\;\;\;\;\;\;\;\;\;\;\;\;\;\;\;\;\;\;
														\;\;\;\;\;\;\;\;\;\;\;\;\;\;\;\;\]
        \IF {$Pe \leq Pe^*-Th$}
                \STATE $Pe\gets Pe^*$
	\ELSE 
	\STATE $SS\gets \textit{SS}  \setminus \textit{S}$
        \ENDIF
\ENDWHILE
\end{algorithmic}
\end{algorithm}

\subsection{Results and analysis }

We apply this algorithm on the hyperspectral image AVIRIS 92AV3C [1], in the same conditions of section 2. 3.

The procedure to construct the estimated reference map \textit{C}$_{est}$ is the same SVM classifier used for classification.

Table. I  shows the results obtained for several thresholds. We can see the effectiveness selection bands of our algorithm, and the important effect of avoiding redundancy.

\begin{table}

\center
\caption{Results illustrate elimination of Redundancy using algorithm based on inequality of Fano, for thresholds ($Th$)}
\begin{tabular}{lllllll}
\hline
\noalign{\smallskip}
${\mathrm Bands }  $& & & & $Th$  \\
${\mathrm retained }$ & 0.00 & 0.001  & 0.008 &0.015 & 0.020 & 0.030 \\
\noalign{\smallskip}

\hline
10 & 55.43 & 55.43 & 55.58 & 53.09 & 60.06 & 71.62 \\
18 & 59.09 & 59.09 & 64.41 & 73.70 & 82.62 & 90.00 \\
20 & 63.08 & 63.08 & 68.50 & 76.15 & 84.36 & - \\
25 & 66.02 & 66.12 & 74.62 & 84.41 & 89.06 & - \\
27 & 69.47 & 69.47 & 76.00 & 86.73 & 91.70 & - \\
30 & 73.54 & 73.54 & 79.04 & 88.68 & - & - \\
35 & 76.06 & 76.06 & 81.38 & 92.36 & - & -\\
40 & 78.96 & 79.41 & 86.48 & - & - & - \\
45 & 80.58 & 80.60 & 89.09 & - & - & - \\
50 & 81.63 & 81.20 & 91.14 & - & - & - \\
53 & 82.27 & 81.22 & 92.67 & - & - & - \\
60 & 86.13 & 86.23 & - & - & - & - \\
70 & 86.97 & 87.55 & - & - & - & -\\
80 & 89.11 & 89.42 & - & - & - & - \\
90 & 90.55 & 90.92 & - & - & - & - \\  
100 & 92.50 & 93.18 & - & - & - & - \\
102 & 92.62 & 93.34 & - & - & - & - \\
114 & 93.34 & - & - & - & - & - \\
\hline
\end{tabular}
\end{table}

Figure.5 shows more detail of the accuracy curves, versus number of bands retained, for several thresholds. This covers all behaviors of the algorithm:\\

\textbf{First:} For the highest threshold values (0.1, 0.05, 0.03 and 0.02) we obtain a hard selection: a few more informative bands are selected; the accuracy of classification is 90\% with less than 20 bands selected.\\
\textbf{Second:} For the medium threshold values (0.015, 0.012, 0.010, 0.008, 0.006), some redundancy is allowed, in order to made increasing the classification accuracy.\\
\textbf{Tired:} For the small threshold values (0.001 and 0), the redundancy allowed becomes useless, we have the same accuracy with more bands.\\
\textbf{Finally:} for the negative thresholds, for example -0.01, we allow all bands to be selected, and we have no action of the algorithm. This corresponds at selection bay ordering bands on mutual information . The performance is low.\\
We can not here that [15] uses two axioms to characterize feature selection. Sufficiency axiom: the subset selected feature must be able to reproduce the training simples without losing information. The necessity axiom "simplest among different alternatives is preferred for prediction". In the algorithm proposed, reducing error probability between the truth map and the estimated minimize the information loosed for the samples training and also the predicate ones.

We note also that we can use the number of features selected like condition to stop the search; so we can obtain an hybrid approach filter-wrapper[16].
\begin{figure}[!h]
\center
\vspace{0.0cm}
\includegraphics[width=3.5in]{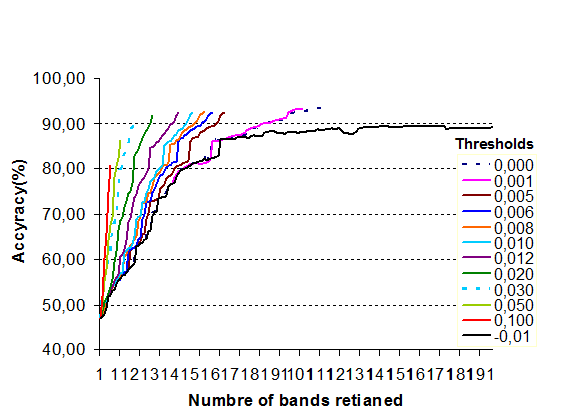}
\caption{Accuracy of classification using the algorithm based on inequality of Fano, using numerous thresholds. }
\end{figure}

\textbf{Partial conclusion:} the algorithm proposed is a very good method to reduce dimensionality of hyperspectral images.

We illustrate in Figure .6, the Ground Truth map originally displayed, like at Figure .1, and the scene classified with our method, for threshold as 0.03, so 18 bands selected.\\ 
Table II indicates the classification accuracy of each class, for several thresholds.\\
\begin{table}
\center
\caption{Accuracy of classification(\%) of each class  for numerous thresholds ($Th$)}
\begin{tabular}[!]{llllllll}
\hline
\noalign{\smallskip}
${\mathrm Class }$&  $Total$& & & & $Th$  \\
    &     ${\mathrm  pixels }$ & 0.00 & 0.001  & 0.008 &0.015 & 0.020 & 0.030 \\
\noalign{\smallskip}
\noalign{\smallskip}
\noalign{\smallskip}
\hline
1  :&        54 & 86.96 & 82.61 & 86.96 & 83.96 & 78.26 & 86.96 \\
2  :&   1434 & 91.07 & 89.40 & 89.54 & 89.12 & 88.01 & 83.96 \\
3  :&     834 & 89.93 & 90.89 & 89.69 & 86.09 & 83.69 & 81.53 \\
4  :&     234 & 96.32 & 83.76 & 86.32 & 87.18 & 87.18 & 86.32 \\
5  : &     597 & 95.93 & 95.53 & 94.34 & 95.93 & 95.93 & 95.53 \\
6  : &     747 & 98.60 & 98.60 & 98.32 & 98.60 & 98.32 & 98.32 \\
7  : &       26 & 84.62 & 84.62 & 84.62 & 84.62 & 84.62 & 84.62 \\
8  : &     489 & 98.37 & 98.37 & 98.78 & 97.96 & 98.78 & 98.78 \\
9  : &       20 & 100 & 100 & 100 & 100 & 100 & 100 \\
10: &     968 & 92.15 & 92.98 & 91.32 & 91.74 & 90.91 & 89.05 \\
11: &   2468 & 93.84 & 94.17 & 92.54 & 92.71 & 91.90 & 91.25 \\
12: &     614 & 91.21 & 93.49 & 92.83 & 92.18 & 88.93 & 87.30 \\
13: &    212 & 98.06 & 98.06 & 98.06 & 98.06 & 98.06 & 98.06 \\
14: & 1294 & 97.53 &  97.86 & 97.22 & 97.84 & 97.99 & 97.53 \\ 
15:&  390 & 79.52 & 77.71 & 75.90 & 74.10 & 78.92 & 64.46 \\
16: &   95 & 93.48 & 93.48 & 93.48 & 93.48 & 93.48 & 93.48 \\

\hline
\end{tabular}
\end{table}

\begin{figure}[!h]
\center
\vspace{0.0cm}
\includegraphics[width=3.5in ]{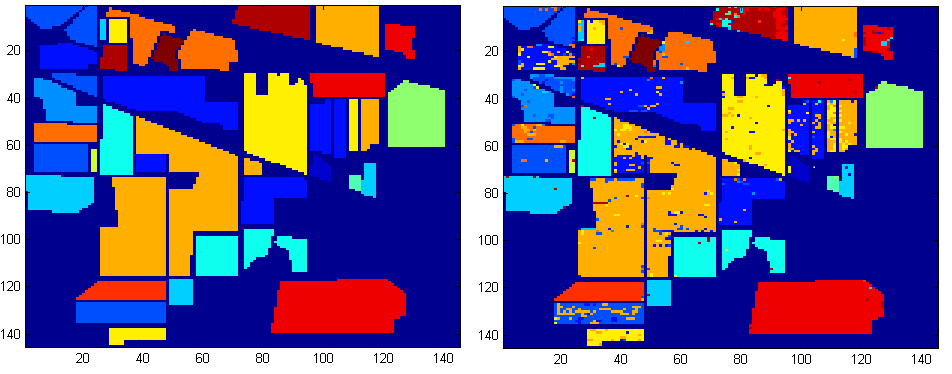}
\caption{Original Grand Truth map(in the left) and the map produced bay our algorithm according to the threshold 0.03 i.e 18 bands (in the right). Acuracy=90\%. }
\end{figure}

\textbf{First :}we can not the effectiveness of this algorithm for particularly the classes with a few number of pixels, for example class number 9. 

\textbf{Second:} we can not that 18 bands (i.e. threshold 0.03) are sufficient to detect materials contained in the region. It’s also shown in Figure .6

\textbf{Tired:} one of important comment is that most of class accuracy change lately when the threshold changes between 0.03 and 0.015

\section{Conclusion }
In this paper we presented the necessity to reduce the number of bands, in classification of Hyperspectral images. Then we have comment results of a filtering redundancy mutual information based scheme. We carried out their effectiveness to select bands able to classify the pixels of ground truth. And also we have carried out their inconvenient as: the elimination of precision by discarding neighboring bands having sensibly the same mutual information with the ground truth map. We introduce an algorithm also based on mutual information and using a measure of error probability (inequality of Fano). To choice a band, it must contributes to reduce error probability. A complementary threshold is added to avoid redundancy. So each band retained has to contribute to reduce error probability by a step equal to threshold even if it caries a redundant information. We can tell that we conserve the useful redundancy by adjusting the complementary threshold. This algorithm is a feature selection methodology. But it’s a wrapper approach, because we use the classifier to make the estimated reference map. This is a limitation that must be avoided by searching another procedure to estimate reference map more rapidly, in order to implement it in a real time applications. This scheme is very interesting to investigate and improve, considering its performance.

\end{document}